\documentclass{article}

\usepackage[preprint]{neurips_2026}

\usepackage[utf8]{inputenc} 
\usepackage[T1]{fontenc}    
\usepackage{hyperref}       
\usepackage{url}            
\usepackage{booktabs}       
\usepackage{amsfonts}       
\usepackage{nicefrac}       
\usepackage{microtype}      
\usepackage{xcolor}         

\usepackage{graphicx,subcaption}
\usepackage{enumitem}
\usepackage{algorithm}
\usepackage{algorithmic}
\usepackage{booktabs}  
\usepackage{ragged2e}
\usepackage{wrapfig}
\usepackage{arydshln}
\usepackage{setspace}
\usepackage{adjustbox} 
\usepackage{framed}

\usepackage{amssymb}
\usepackage{mathtools}
\usepackage{amsthm}

\newbool{makeBlue}
\setbool{makeBlue}{false} 
\newcommand{\conditionalblue}[1]{%
  \ifbool{makeBlue}{%
    \textcolor{blue}{#1}%
  }{%
    #1%
  }%
}

\theoremstyle{plain}

\theoremstyle{definition}

\theoremstyle{remark}

\title{MSCR: Jointly Balancing Utilization Across Modalities and Discovering Synergistic Information}

\author{%
  \textbf{Xinyu Chen$^{1}$, Liangjian Wen$^{1,7\dagger}$, Jiang Duan$^{1,8}$, Dongkai Wang$^{1}$, Yong Dai$^{3}$,}\\
  \textbf{Jiayu Bai$^{1}$, Jianzhuang Liu$^{2}$, Zhen Tian$^{5}$, Guoping Qiu$^{6}$, Zhao Kang$^{4}$}\\[0.5em]
  $^{1}$School of Computing and Artificial Intelligence, Southwestern University of Finance and \\ Economics\\
  $^{2}$Shenzhen Institutes of Advanced Technology, Chinese Academy of Sciences\\
  $^{3}$X-Humanoid\\
  $^{4}$University of Electronic Science and Technology of China\\
  $^{5}$Guangdong Laboratory of Artificial Intelligence and Digital Economy\\
  $^{6}$Computer Science, University of Nottingham\\
  $^{7}$Artificial Intelligence and Digital Finance Key Laboratory of Sichuan Province\\
  $^{8}$Chengdu Everimaging Science and Technology Co., Ltd\\[0.5em]
  \texttt{{chenxinyu991}@gmail.com}, \texttt{wlj6816@gmail.com}
}

\begin{document}

\maketitle

\begingroup
\renewcommand{\thefootnote}{\fnsymbol{footnote}}
\footnotetext[2]{Corresponding author.}
\endgroup

\begin{abstract}
In multimodal learning, modality imbalance often lets one modality dominate optimization and keeps the model from fully exploiting cross-modal interactions. Balance-oriented methods such as the Multimodal Competition Regularizer (MCR) perform well by alleviating this imbalance during training. These dynamic balancing methods, however, mainly adjust modality contributions and do not consider how imbalance affects the formation of synergistic information during fusion. Synergy is a major source of multimodal advantage, because it is predictive gain that cannot be recovered from unimodal evidence alone. We propose the Multimodal Synergistic Competition Regularizer (MSCR), a fusion method that jointly promotes balanced modality utilization and the discovery of synergistic information. MSCR has two complementary objectives. {An anti-dominance objective penalizes large modality-perturbation sensitivities to regularize modality reliance, with the aim of leaving ``representation room'' for joint predictive cues. A synergy objective encourages the fused predictor to outperform an additive unimodal reference.} An adaptive gate further concentrates the regularization on samples that need stronger collaborative modeling. Experiments on five benchmarks show that MSCR consistently improves performance over MCR and strengthens synergistic capability. It raises conditional recovery, the accuracy of the fused prediction on samples that the unimodal predictors misclassify, as well as the share of the model's output that InterSHAP attributes to cross-modal interactions. Its gain in conditional recovery persists when the fusion module is Transformer-based. Code is available at \url{https://github.com/chenxinyu991-collab/MSCR}.
\end{abstract}

\section{Introduction}
\label{sec:introduction}
Multimodal learning integrates heterogeneous signals such as vision, audio, and language, and has achieved strong performance~\cite{liang2024foundations,tsai2019multimodal,zadeh2017tensor}. Recent work has extended it to high-resolution visual understanding~\cite{mao2026pay} and knowledge-driven reasoning for web agents~\cite{guo2026webcogreasoner}. Its main advantage comes from combining complementary information. Jointly trained multimodal models, however, often suffer from modality imbalance: the model relies too heavily on a dominant modality, which hinders cross-modal fusion~\cite{wang2020makes,yao2022modality,huang2022modality}. Prior work has explored several training strategies to alleviate this imbalance, including suppressing information from the dominant modality~\cite{wu2022characterizing,du2023unimodal,wei2024mmpareto}, monitoring modality contributions to adaptively modulate gradients~\cite{sun2021learning,peng2022balanced,li2023boosting,kontras2024improving}, and decomposing multimodal learning into alternating unimodal optimization~\cite{zhang2024multimodal}, which preserves cross-modal interaction while reducing biased optimization dynamics.

The Multimodal Competition Regularizer (MCR)~\cite{kontras2025balancing} was recently proposed to mitigate modality imbalance explicitly. {MCR perturbs modality features to characterize each modality's contribution to the joint prediction. It treats each encoder as a greedy player that maximizes its own importance, possibly at the expense of the others.} This competitive game balances modality utilization well, but it does not address the discovery of synergistic information~\cite{williams2010nonnegative,bertschinger2014quantifying}, that is, task information that is available only when the modalities are observed jointly. Synergistic gain comes from collaborative interactions between modalities, so forcing modalities to compete does not by itself push the model to find information that no single modality can provide. More broadly, earlier methods and MCR alike focus on adjusting modality contributions during training and pay little attention to how imbalance and purely competitive dynamics affect the formation of synergistic information~\cite{xiu2026comprehensive, wen2025infmasking}.

Synergistic information is a major source of multimodal advantage~\cite{wen2025infmasking,xin2025i2moe,wen2026dependency}, because it provides predictive gain beyond unimodal evidence. Even a strong modality can be insufficient: identifying sarcasm, for example, requires interpreting facial expression and vocal intonation together~\cite{zadeh2016mosi,zadeh2018multimodal}. A model that over-relies on a dominant modality tends to exploit only the information that modality can supply on its own (its unique and redundant components) and to overlook synergistic information that requires joint observation.

We therefore introduce the Multimodal Synergistic Competition Regularizer (MSCR), which combines modality balancing with explicit synergy discovery. MSCR first limits the dominance of any single modality with an anti-dominance objective. It perturbs modality features in the latent space to measure how strongly the fused prediction depends on each modality, and penalizes the network for relying too heavily on any one of them. {Discouraging excessive modality reliance may leave ``representation room'' for cross-modal interactions. This is the motivation for the constraint, but reducing perturbation sensitivity alone does not guarantee synergy. MSCR therefore adds a synergy objective based on an additive reference formed from independent unimodal predictions.} Requiring the fused model to achieve a discriminative gain over this naive ensemble encourages the network to exploit synergistic information. Finally, to keep optimization stable and avoid over-regularization, an instance-level proficiency-aware gate tracks unimodal explanatory power during training and scales the MSCR regularizer for each sample, so that it acts mainly on difficult samples on which unimodal evidence is insufficient.

Our main contributions are as follows:
\begin{itemize}[leftmargin=*, itemsep=0.6em]
    \item We point out that modality imbalance suppresses the discovery of synergistic information, and we propose MSCR, a fusion method that jointly optimizes balanced modality utilization and synergistic gain.
    \item Motivated by an information-theoretic view, MSCR limits single-modality dominance in the fusion network and explicitly rewards joint predictions that outperform an additive unimodal reference. A proficiency-aware adaptive gate applies this regularization mainly to samples that unimodal evidence cannot resolve, which keeps optimization stable.
    \item Experiments show that MSCR consistently improves downstream performance and synergistic capability, as measured by conditional recovery and an independent InterSHAP~\cite{wenderoth2025measuring} analysis, and that its advantage in conditional recovery persists with a Transformer-based fusion module.
\end{itemize}

\section{Preliminaries: Multimodal Competition Regularizer}
\label{sec:preliminaries}

To mitigate modality imbalance, MCR~\cite{kontras2025balancing} regulates the information contributed by each modality. Following MCR, we express the joint mutual information through an information-theoretic identity:
\begin{equation}
\label{eq:mcr_mi_decomposition}
I(X_1,X_2;Y)
=
\underbrace{I(X_1;Y \mid X_2) + I(X_2;Y \mid X_1)}_{\text{{Conditional Task-Relevant Information}}}
+
\underbrace{I(X_1;X_2)}_{\text{Shared Info}}
-
\underbrace{I(X_1;X_2 \mid Y)}_{\text{{Conditional Shared Info}}} .
\end{equation}
Based on this decomposition, MCR uses three regularization terms:
\begin{equation}
\label{eq:mcr_loss}
\mathcal{L}_{\mathrm{MCR}}
=
\mathcal{L}_{\mathrm{MIPD}}
+
\mathcal{L}_{\mathrm{Con}}
+
\mathcal{L}_{\mathrm{CEB}} .
\end{equation}

\paragraph{{Modality Contribution Learning} $(\mathcal{L}_{\mathrm{MIPD}})$.}
{MCR introduces the Mutual Information Perturbed Difference (MIPD) to characterize each modality's contribution to the joint prediction, motivated by the conditional mutual information $I(X_i;Y \mid X_j)$ for $i\neq j$. Under partial information decomposition (PID), this quantity includes both information unique to modality $i$ and synergistic information available only through joint observation. MIPD uses changes in the prediction under modality perturbation as a practical proxy for contribution. Randomly permuting the latent representations of modality 1 across a mini-batch preserves their empirical marginal distribution but breaks their sample-wise pairing with modality 2 and the labels. Writing the perturbed modality as $\widetilde{X}_1$, MCR uses the negative Jensen--Shannon (JS) divergence~\cite{lin1991divergence} between the original and perturbed joint predictions as a stable optimization surrogate:}
\begin{equation}
\label{eq:mcr_mipd}
\mathcal{L}_{\mathrm{MIPD}_1}
=
-\mathbb{E}
\left[
\mathrm{JSD}
\left(
p(y \mid x_2,x_1),
p(y \mid x_2,\widetilde{x}_1)
\right)
\right].
\end{equation}
$\mathcal{L}_{\mathrm{MIPD}_2}$ is defined analogously by perturbing $X_2$. {Minimizing $\mathcal{L}_{\mathrm{MIPD}_i}$ maximizes this prediction discrepancy. Together with task supervision, it encourages the corresponding encoder to learn features that are used, rather than ignored, by the fusion network in the joint prediction.}

\paragraph{Shared Information Alignment $(\mathcal{L}_{\mathrm{Con}})$.}
{MCR uses a supervised cross-modal contrastive loss~\cite{khosla2020supervised} to encourage shared class-relevant structure. Positive pairs include cross-modal representations from samples with the same class label, not just the two modalities of a single sample. The loss encourages greater agreement within these positive pairs than with other samples in the mini-batch.}

\paragraph{Irrelevant Information Filtering $(\mathcal{L}_{\mathrm{CEB}})$.}
{MCR interprets the conditional cross-modal dependence $I(X_1;X_2 \mid Y)$ as task-irrelevant shared information. It regularizes this dependence with a penalty inspired by the Conditional Entropy Bottleneck (CEB)~\cite{fischer2020conditional}. A small reconstruction head $h$ predicts the concatenated modality representations from the label $Y$. Minimizing the reconstruction error}
\begin{equation}
\label{eq:mcr_ceb}
\mathcal{L}_{\mathrm{CEB}}
=
\mathbb{E}
\left[
\left\|
[f_1(x_1), f_2(x_2)] - h(y)
\right\|^2
\right],
\end{equation}
{encourages representations that can be predicted from the task label and discourages variation that the label does not explain. This penalty is a practical regularizer, not an exact measure of $I(X_1;X_2 \mid Y)$.}

\paragraph{The Game of Competition.}
{MCR treats each modality encoder as a player in a competitive game. Under the default greedy strategy ($k=-1$), encoder $i$ minimizes its own $\mathcal{L}_{\mathrm{MIPD}_i}$ while maximizing the other modality's $\mathcal{L}_{\mathrm{MIPD}_j}$. The MIPD gradient used to update encoder parameters $\theta_i$ by gradient descent is proportional to}
\begingroup
\begin{equation*}
\nabla_{\theta_i}\mathcal{L}_{\mathrm{MIPD}_i}
-
\nabla_{\theta_i}\mathcal{L}_{\mathrm{MIPD}_j},
\qquad i,j\in\{1,2\},\quad i\neq j.
\end{equation*}
\endgroup
{Each loss is a negative prediction discrepancy, so this update encourages each encoder to increase its own influence on the joint prediction while reducing the other's. The competition comes from these encoder-specific gradient rules, not from simply minimizing the sum of the two MIPD losses. It is intended to mitigate modality imbalance, but it neither explicitly isolates nor directly optimizes synergistic information.}

\section{Method}
\label{sec:method}
\subsection{The General Framework}
\label{sec:framework}

We consider a supervised multimodal learning task. For simplicity of exposition, we describe the method with two input modalities $X_1$ and $X_2$ and a target label $Y$; the framework extends to more than two modalities. During training, for a batch of $B$ samples $\{(\mathbf{x}_1^{(b)}, \mathbf{x}_2^{(b)}, y^{(b)})\}_{b=1}^B$, two modality-specific encoders $f_1$ and $f_2$ map the inputs into latent representations $\mathbf{z}_1 = f_1(\mathbf{x}_1) \in \mathbb{R}^{B \times d_1}$ and $\mathbf{z}_2 = f_2(\mathbf{x}_2) \in \mathbb{R}^{B \times d_2}$. A fusion network $f_c$ combines these representations into the joint prediction $p_{\mathrm{joint}} = f_c(\mathbf{z}_1, \mathbf{z}_2)$, and two auxiliary unimodal predictors $q_1$ and $q_2$ produce the modality-specific outputs $p_1 = q_1(\mathbf{z}_1)$ and $p_2 = q_2(\mathbf{z}_2)$. For classification, $p_{\mathrm{joint}}$, $p_1$, and $p_2$ are predictive distributions over the classes; Appendix~\ref{app:experimental_details} gives the regression variant.

Partial information decomposition (PID)~\cite{williams2010nonnegative,bertschinger2014quantifying} splits the task-relevant information $I(X_1, X_2;Y)$ into redundant information $(R)$ that either modality provides on its own, unique information $(U_1, U_2)$ specific to one modality, and synergistic information $(S)$ that emerges only when the modalities are combined. {$\mathcal{L}_{\mathrm{MCR}}$ regulates modality contributions through competition, but it does not explicitly separate or directly target synergistic information $S$, which depends on inter-modal interactions.}

MSCR addresses this gap as a lightweight regularizer that acts only on the fusion network $f_c$ and does not interfere with the modality-specific encoders $f_1$ and $f_2$. {It constrains modality-perturbation sensitivity with the aim of mitigating excessive modality reliance and leaving ``representation room'' for joint predictive cues. A separate objective explicitly encourages joint predictive gain beyond unimodal aggregation.} The overall training objective is
\begin{equation}
\label{eq:total_loss}
\mathcal{L}_{\mathrm{total}}
=
\mathcal{L}_{\mathrm{task}}
+
\mathcal{L}_{\mathrm{MCR}}
+
\mathcal{L}_{\mathrm{MSCR}},
\end{equation}
where $\mathcal{L}_{\mathrm{task}}$ is the supervised task loss computed on the joint prediction and on the two unimodal predictions, $\mathcal{L}_{\mathrm{MCR}}$ balances modality contributions as described in Section~\ref{sec:preliminaries}, and $\mathcal{L}_{\mathrm{MSCR}}$ is our synergy-enhancing regularizer:
\begin{equation}
\label{eq:mscr_loss}
\mathcal{L}_{\mathrm{MSCR}}
=
g \cdot
\left(
\lambda_{\mathrm{dom}}\mathcal{L}_{\mathrm{dom}}
+
\lambda_{\mathrm{syn}}\mathcal{L}_{\mathrm{syn}}
\right).
\end{equation}
{The anti-dominance objective $\mathcal{L}_{\mathrm{dom}}$ penalizes large modality-perturbation sensitivities in the fusion network, placing the most weight on the largest sensitivity. The synergy objective $\mathcal{L}_{\mathrm{syn}}$ encourages joint predictive gain beyond unimodal aggregation.} The proficiency-aware gate $g \in (0,1)$ scales both objectives according to sample difficulty. The gate and both objectives are computed per sample, and $\mathcal{L}_{\mathrm{MSCR}}$ is averaged over the mini-batch. MSCR is switched on after an initial training stage and applied within a fixed interval of epochs (Appendix~\ref{app:experimental_details}); Algorithm~\ref{alg:mscr} in Appendix~\ref{app:pseudo_code} gives the complete training procedure.

The regularizers operate on distinct parameter groups. For MCR, the fusion parameters are held fixed when computing MIPD, while gradients with respect to the fusion network's inputs are retained to update the modality encoders; the CEB term also updates its reconstruction head. For MSCR, modality representations are detached in all joint and perturbed predictions, and the unimodal reference and gate are also detached, so its gradients update only the fusion network. MIPD and $\mathcal{L}_{\mathrm{dom}}$ therefore act on different parameters. These restrictions apply only to the regularizers; $\mathcal{L}_{\mathrm{task}}$ trains the model through the standard gradient paths.
\begin{figure}[t]
    \centering
    \includegraphics[width=\linewidth]{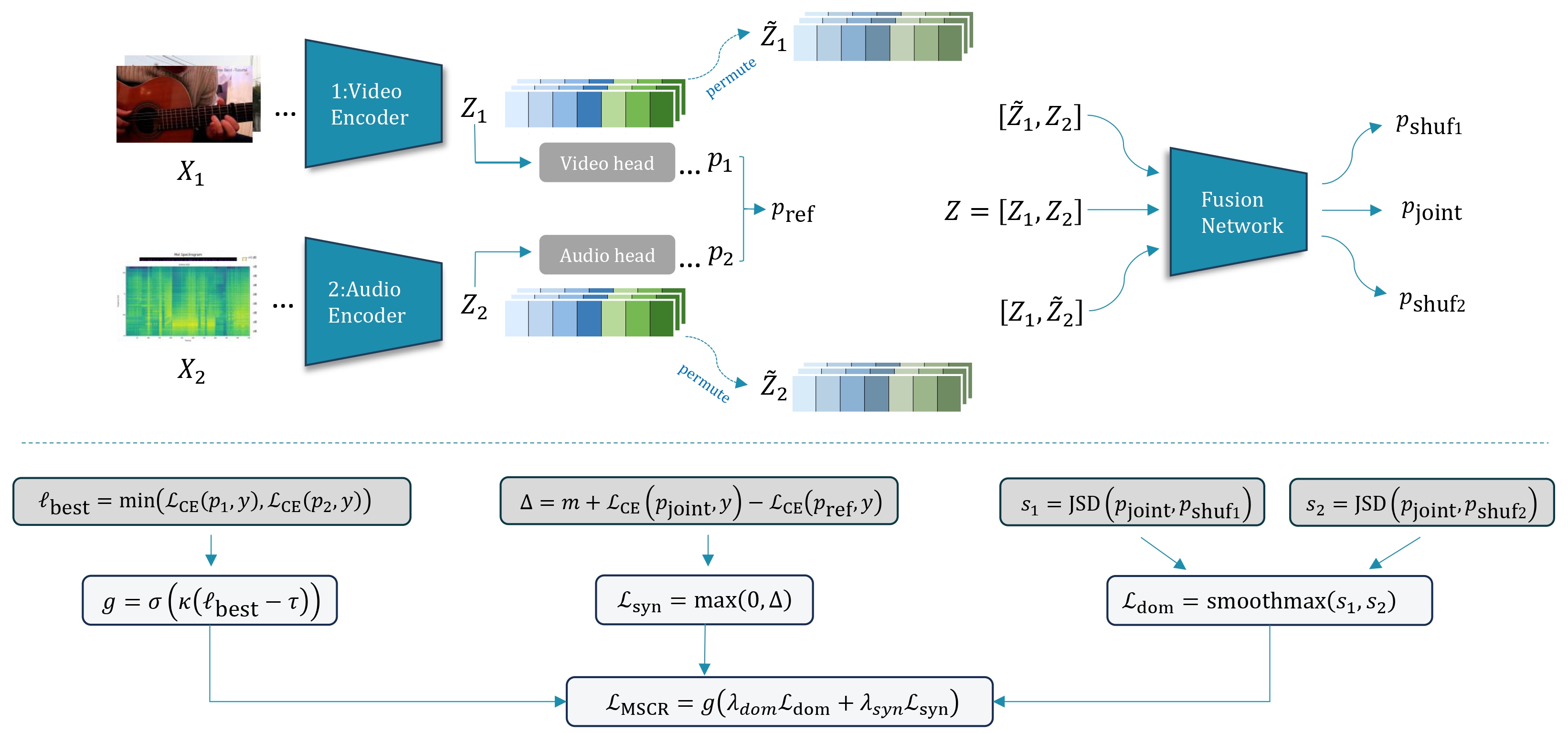}
    \caption{\textbf{Multimodal Synergistic Competition Regularizer} (MSCR)\textbf{:} MSCR suppresses modality dominance and encourages the discovery of synergistic information. Raw data ($X_1, X_2$) are encoded into representations ($Z_1, Z_2$). These and their permuted variants ($\tilde{Z}_1, \tilde{Z}_2$) are fed into the fusion network to produce joint ($p_{\text{joint}}$) and shuffled ($p_{\text{shuf1}}, p_{\text{shuf2}}$) predictions, while the unimodal heads provide an additive reference ($p_{\text{ref}}$). The MSCR loss is built from three mechanisms: {$\mathcal{L}_{\text{dom}}$ uses a smoothmax to penalize modality-perturbation sensitivities, placing the most weight on the largest sensitivity;} $\mathcal{L}_{\text{syn}}$ requires the joint prediction to outperform the unimodal reference by a margin $m$, with $\Delta$ denoting the hinge argument; and a proficiency-aware gate ($g$) scales the regularizer according to the best unimodal loss ($\ell_{\text{best}}$).}
    \label{fig:mscr_framework}
    \vspace*{-4mm}
\end{figure}

\subsection{Anti-Dominance Objective}
\label{sec:anti_dominance}

{Modality dominance occurs when the fusion network relies too heavily on readily accessible cues from one modality and overlooks the others, a common problem in multimodal learning. MIPD encourages each encoder to learn features that contribute to the joint prediction, but their availability does not ensure that the fusion network uses them jointly. We therefore regularize the strength of the fused prediction's dependence on individual modalities. The aim is to mitigate excessive modality reliance and leave ``representation room'' for joint predictive cues.}

{We use the MIPD-based perturbation measure from Section~\ref{sec:preliminaries}, but reverse its optimization direction when updating the fusion network. Permuting modality $i$ across a mini-batch gives the perturbed predictions $p_{\mathrm{shuf}1}=f_c(\widetilde{\mathbf{z}}_1,\mathbf{z}_2)$ and $p_{\mathrm{shuf}2}=f_c(\mathbf{z}_1,\widetilde{\mathbf{z}}_2)$. For each sample, the modality-perturbation sensitivity is}
\begingroup
\begin{equation}
\label{eq:jsd_sensitivity}
s_i = \operatorname{JSD}\left(p_{\mathrm{joint}} \parallel p_{\mathrm{shuf}i}\right) = -\ell_{\mathrm{MIPD}_i},
\qquad i \in \{1,2\},
\end{equation}
\endgroup
{where $\ell_{\mathrm{MIPD}_i}$ is the per-sample loss term in Eq.~\eqref{eq:mcr_mipd}. A larger $s_i$ means that the prediction changes more when modality $i$ is replaced with an unpaired representation. This sensitivity is a proxy for modality reliance, not a measure of purely unimodal information. MIPD encourages an encoder to increase its own contribution; our objective instead penalizes large sensitivities when updating the fusion network. We combine these sensitivities with a differentiable smooth maximum:}
\begin{equation}
\label{eq:mscr_dom}
\mathcal{L}_{\text{dom}} = \operatorname{smoothmax}(s_1, s_2) = \frac{1}{\tau_{\text{dom}}} \log \left( \exp({\tau_{\text{dom}} s_1}) + \exp({\tau_{\text{dom}} s_2}) \right),
\end{equation}
{where $\tau_{\text{dom}} > 0$ controls the sharpness of the approximation. Since $\partial\mathcal{L}_{\mathrm{dom}}/\partial s_i = \exp(\tau_{\mathrm{dom}}s_i)/\sum_j\exp(\tau_{\mathrm{dom}}s_j)$, larger sensitivities receive more weight in the penalty. The most heavily penalized modality can therefore vary from sample to sample. We detach modality representations in this branch, so the objective changes how the fusion network uses the available features without directly updating the encoders.}

{Reducing perturbation sensitivity alone does not guarantee synergy, because permuting a modality can also disrupt useful cross-modal interactions. Leaving ``representation room'' is therefore a design motivation, not a guaranteed redistribution of information. We use $\mathcal{L}_{\mathrm{dom}}$ as an auxiliary constraint together with the task loss, which encourages accurate predictions, and $\mathcal{L}_{\mathrm{syn}}$, which encourages joint predictive gain beyond unimodal aggregation.}

\subsection{Synergy Objective}
\label{sec:synergy_objective}

Even when no single modality dominates, the joint prediction may be no better than an aggregate of unimodal evidence and miss the predictive gains that arise only from combining modalities. MSCR therefore adds a margin-based synergy objective that rewards joint predictive gain beyond unimodal aggregation.

As a simple proxy for synergistic capability, we define an additive unimodal reference distribution $p_{\text{ref}} = \frac{1}{2}(p_1 + p_2)$, where $p_1$ and $p_2$ are the outputs of the auxiliary unimodal predictors $q_1$ and $q_2$ of the model being trained. This reference simply averages independent unimodal decisions. A fusion network that extracts meaningful synergistic information should substantially outperform this naive ensemble.

The task losses of the joint prediction and the additive reference are
\begin{equation}
\ell_{\text{joint}} = \mathcal{L}_{\text{CE}}(p_{\text{joint}}, y), \quad \text{and} \quad \ell_{\text{ref}} = \mathcal{L}_{\text{CE}}(p_{\text{ref}}, y),
\end{equation}
where $y$ is the ground-truth label and $\mathcal{L}_{\text{CE}}(p, y) = -\log p(y)$ is the cross-entropy loss. Thus $\ell_{\text{ref}}$ is the negative log of the average probability that the two unimodal predictors assign to the correct class. The synergy objective is
\begin{equation}
\label{eq:mscr_syn}
\mathcal{L}_{\text{syn}} = \max(0,\, m + \ell_{\text{joint}} - \ell_{\text{ref}}),
\end{equation}
where $m > 0$ is a fixed margin. Minimizing $\mathcal{L}_{\text{syn}}$ drives the model toward $\ell_{\text{joint}} \leq \ell_{\text{ref}} - m$, so the joint predictor must reach a lower loss than the additive baseline. A sample that satisfies this condition no longer contributes to $\mathcal{L}_{\text{syn}}$, and because $\ell_{\text{ref}}$ is detached, the objective can be reduced only by improving the joint prediction. {The objectives complement each other: $\mathcal{L}_{\text{dom}}$ regularizes modality reliance through perturbation sensitivity, while $\mathcal{L}_{\text{syn}}$ explicitly encourages joint predictive gain beyond the unimodal reference.}

\subsection{Proficiency-Aware Adaptive Gating}
\label{sec:adaptive_gating}

Samples differ in how much regularization they need. When at least one unimodal predictor already classifies a sample correctly with high confidence, strong regularization on that sample may be unnecessary. When both unimodal predictions are ambiguous or wrong, the fusion model has to mine synergistic information to reduce task uncertainty. MSCR therefore uses an instance-level, proficiency-aware gate to adjust the strength of $\mathcal{L}_{\text{MSCR}}$.

For a given sample, let the unimodal task losses be $\ell_1 = \mathcal{L}_{\text{CE}}(p_1, y)$ and $\ell_2 = \mathcal{L}_{\text{CE}}(p_2, y)$. We measure the best unimodal proficiency by the minimum unimodal loss, $\ell_{\text{best}} = \min(\ell_1, \ell_2)$, and compute the gate $g \in (0, 1)$ with a smooth step function:
\begin{equation}
\label{eq:mscr_gate}
g = \sigma\bigl(\kappa(\ell_{\text{best}} - \tau)\bigr),
\end{equation}
where $\sigma(\cdot)$ is the sigmoid function, and $\kappa > 0$ and $\tau > 0$ set the sharpness and the threshold of the gate ($\tau$ is distinct from the smoothmax temperature $\tau_{\text{dom}}$). For classification, $\ell_{\text{best}} > \tau$ means that even the more confident unimodal predictor assigns a probability below $e^{-\tau}$ to the correct class, which is about $0.30$ and $0.22$ for the thresholds $\tau = 1.2$ and $\tau = 1.5$ used in our experiments.

The gate works as a soft switch. It opens ($g \to 1$) when unimodal evidence is insufficient ($\ell_{\text{best}} > \tau$), so both objectives concentrate on hard samples, and it closes ($g \to 0$) for easy samples that a single modality already handles ($\ell_{\text{best}} \ll \tau$), so that on these samples the fusion network is trained almost entirely by the task loss.

For the trimodal settings, $p_{\text{ref}}$ averages the three unimodal predictions, the smooth maximum in Eq.~\eqref{eq:mscr_dom} is taken over the three sensitivities, and $\ell_{\text{best}}$ is the smallest of the three unimodal losses.

\section{Experiments}
\label{sec:experiments}
\subsection{Main Results on Real-World Benchmarks}

We first evaluate MSCR on real-world multimodal benchmarks. Following the evaluation protocol of MCR~\cite{kontras2025balancing}, we use CREMA-D~\cite{cao2014cremad} for emotion recognition, AVE~\cite{tian2018audio} for event recognition, UCF101~\cite{soomro2012ucf101} for action recognition, and CMU-MOSI~\cite{zadeh2016mosi} and CMU-MOSEI~\cite{zadeh2018multimodal} for multimodal sentiment and emotion analysis. The datasets cover bimodal and trimodal settings as well as classification and regression-style tasks. We report prior multimodal learning methods under the same protocol and focus on the comparison between MSCR and the state-of-the-art baseline MCR.

Table~\ref{table:main_results} also includes the training strategies evaluated by MCR. Unimodals reports each modality trained on its own, and Ensemble combines these unimodal predictions without further training. Joint Training optimizes all modalities under a single multimodal loss, and Multi-Loss~\cite{vielzeuf2018centralnet} adds unimodal task losses. Uni-Pre Frozen and Uni-Pre Finetuned start from pre-trained unimodal encoders that are kept frozen or fine-tuned jointly, respectively. The remaining baselines are the balancing methods reviewed in Section~\ref{sec:related_work}. In the table header, the second row gives the unimodal encoder and the third the modalities used (V: video, A: audio, T: text), and UCF denotes UCF101. Entries are the mean $\pm$ standard deviation over three folds (CREMA-D, UCF101) or three random seeds (AVE, MOSI, MOSEI), and the best result in each column is in bold. The encoders, fusion network, and training schedule follow MCR; Appendices~\ref{app:experimental_details} and~\ref{app:dataset_backbone_details} give the details.

\begin{table}[!ht]
\vspace{-2mm}
\captionsetup{position=above, skip=5pt} 
\caption{Comparison of MSCR with prior multimodal training methods on five real-world benchmarks, covering bimodal and trimodal settings. Following MCR, the regression-style tasks (CMU-MOSI and CMU-MOSEI) are evaluated by binary accuracy, and the other datasets are standard classification tasks. For a fair comparison, we follow the evaluation protocol of MCR. MCR results are reproduced by us, while the remaining baseline results are taken from that work.}
\label{table:main_results}
\centering
\resizebox{\textwidth}{!}{
\begin{tabular}{lccccccccc}
\toprule[1.5pt]
& \multicolumn{2}{c}{CREMA-D} & \multicolumn{2}{c}{AVE} & \multicolumn{1}{c}{UCF} & \multicolumn{2}{c}{MOSI} & \multicolumn{2}{c}{MOSEI} \\
& \multicolumn{1}{c}{ResNet} & \multicolumn{1}{c}{Conformer} & \multicolumn{1}{c}{ResNet} & \multicolumn{1}{c}{Conformer} & \multicolumn{1}{c}{ResNet} & \multicolumn{4}{c}{Transformer} \\
Method & V-A & V-A & V-A & V-A & V-A & V-T & V-A-T & V-T & V-A-T \\ \midrule[1.2pt]

Unimodals
& \multicolumn{1}{c}{\begin{tabular}{c} V: 55.4{\tiny$\pm$3.0}\\ A: 60.6{\tiny$\pm$2.3} \end{tabular}}
& \multicolumn{1}{c}{\begin{tabular}{c} V: 69.4{\tiny$\pm$2.0}\\ A: 76.0{\tiny$\pm$2.6} \end{tabular}}
& \multicolumn{1}{c}{\begin{tabular}{c} V: 45.7{\tiny$\pm$1.6}\\ A: 62.6{\tiny$\pm$0.9} \end{tabular}}
& \multicolumn{1}{c}{\begin{tabular}{c} V: 75.5{\tiny$\pm$1.2}\\ A: 76.5{\tiny$\pm$2.4} \end{tabular}}
& \multicolumn{1}{c}{\begin{tabular}{c} V: 38.5{\tiny$\pm$0.9}\\ A: 30.3{\tiny$\pm$1.5} \end{tabular}}
& \multicolumn{2}{c}{\begin{tabular}{c} V: 54.1{\tiny$\pm$3.7}\\ A: 53.7{\tiny$\pm$0.6}\\ T: 72.1{\tiny$\pm$3.3} \end{tabular}}
& \multicolumn{2}{c}{\begin{tabular}{c} V: 64.8{\tiny$\pm$0.2}\\ A: 64.4{\tiny$\pm$0.2}\\ T: 78.9{\tiny$\pm$1.7} \end{tabular}} \\
\midrule

Ensemble
& 71.7{\tiny$\pm$2.2} & 84.6{\tiny$\pm$1.0} & 70.5{\tiny$\pm$0.2} & 88.4{\tiny$\pm$2.2} & 52.8{\tiny$\pm$0.5} & 70.5{\tiny$\pm$2.1} & 67.2{\tiny$\pm$1.5} & 78.4{\tiny$\pm$0.7} & 77.2{\tiny$\pm$0.6} \\

Joint Training
& 62.6{\tiny$\pm$5.8} & 74.6{\tiny$\pm$2.2} & 66.7{\tiny$\pm$1.5} & 82.2{\tiny$\pm$0.8} & 47.7{\tiny$\pm$1.5} & 73.0{\tiny$\pm$1.3} & 73.6{\tiny$\pm$1.3} & 80.5{\tiny$\pm$0.2} & 80.8{\tiny$\pm$0.3} \\

Multi-Loss~\cite{vielzeuf2018centralnet}
& 69.2{\tiny$\pm$1.8} & 82.6{\tiny$\pm$0.9} & 70.1{\tiny$\pm$0.9} & 86.3{\tiny$\pm$1.1} & 51.1{\tiny$\pm$1.8} & 72.1{\tiny$\pm$0.4} & 73.6{\tiny$\pm$2.9} & 80.0{\tiny$\pm$0.7} & 80.2{\tiny$\pm$0.5} \\

Uni-Pre Frozen
& 72.4{\tiny$\pm$1.8} & 85.0{\tiny$\pm$1.8} & 72.2{\tiny$\pm$0.3} & 87.2{\tiny$\pm$2.4} & 53.0{\tiny$\pm$0.9} & 73.3{\tiny$\pm$1.8} & 72.7{\tiny$\pm$1.6} & 79.9{\tiny$\pm$0.5} & 79.8{\tiny$\pm$0.3} \\

Uni-Pre Finetuned
& 73.3{\tiny$\pm$1.8} & 82.4{\tiny$\pm$2.0} & 72.5{\tiny$\pm$1.3} & 86.5{\tiny$\pm$0.8} & 53.5{\tiny$\pm$1.3} & 73.1{\tiny$\pm$2.3} & 73.7{\tiny$\pm$0.7} & 80.3{\tiny$\pm$0.4} & 80.3{\tiny$\pm$0.2} \\

MSLR~\cite{yao2022modality}
& 56.5{\tiny$\pm$2.4} & 77.1{\tiny$\pm$2.4} & 67.3{\tiny$\pm$2.2} & 81.0{\tiny$\pm$1.4} & 50.9{\tiny$\pm$3.9} & $\times$ & $\times$ & $\times$ & $\times$ \\

MMCosine~\cite{xu2023mmcosine}
& 59.3{\tiny$\pm$1.5} & 74.0{\tiny$\pm$0.3} & 65.0{\tiny$\pm$1.4} & 83.8{\tiny$\pm$0.8} & 47.3{\tiny$\pm$4.1} & $\times$ & $\times$ & $\times$ & $\times$ \\

OGM~\cite{peng2022balanced}
& 65.6{\tiny$\pm$3.8} & 82.4{\tiny$\pm$1.0} & 67.3{\tiny$\pm$0.6} & 79.7{\tiny$\pm$1.4} & 51.8{\tiny$\pm$1.9} & 73.9{\tiny$\pm$1.1} & $\otimes$ & 79.7{\tiny$\pm$0.6} & $\otimes$ \\

AGM~\cite{li2023boosting}
& 69.3{\tiny$\pm$1.4} & 78.5{\tiny$\pm$1.6} & 68.4{\tiny$\pm$1.1} & 85.3{\tiny$\pm$0.5} & 51.0{\tiny$\pm$1.6} & 74.0{\tiny$\pm$1.3} & 73.9{\tiny$\pm$1.9} & 79.3{\tiny$\pm$0.4} & 80.2{\tiny$\pm$0.3} \\

MLB~\cite{kontras2024improving}
& 71.9{\tiny$\pm$2.2} & 85.2{\tiny$\pm$0.9} & 71.6{\tiny$\pm$0.2} & 86.7{\tiny$\pm$0.3} & 52.2{\tiny$\pm$1.7} & 72.4{\tiny$\pm$1.7} & 74.2{\tiny$\pm$1.7} & 80.1{\tiny$\pm$0.5} & 80.5{\tiny$\pm$0.4} \\

ReconBoost~\cite{hua2024reconboost}
& 69.0{\tiny$\pm$2.4} & 84.8{\tiny$\pm$1.8} & 68.4{\tiny$\pm$1.7} & 86.1{\tiny$\pm$0.7} & 50.2{\tiny$\pm$4.0} & $\times$ & $\times$ & $\times$ & $\times$ \\

MMPareto~\cite{wei2024mmpareto}
& 69.0{\tiny$\pm$2.5} & 83.8{\tiny$\pm$0.8} & 73.0{\tiny$\pm$1.3} & 87.4{\tiny$\pm$1.3} & 51.4{\tiny$\pm$2.2} & 73.4{\tiny$\pm$1.0} & 73.7{\tiny$\pm$0.6} & 79.3{\tiny$\pm$0.6} & 79.5{\tiny$\pm$0.8} \\

D\&R~\cite{wei2024diagnosing}
& 70.6{\tiny$\pm$1.3} & 85.0{\tiny$\pm$0.4} & 72.3{\tiny$\pm$1.5} & \textbf{91.0{\tiny$\pm$0.7}} & 49.3{\tiny$\pm$1.0} & $\times$ & $\times$ & $\times$ & $\times$ \\

\hdashline
MCR
& 74.5{\tiny$\pm$0.8}
& 85.1{\tiny$\pm$0.5}
& 73.5{\tiny$\pm$0.0}
& 87.4{\tiny$\pm$0.2}
& 52.9{\tiny$\pm$2.4}
& 75.3{\tiny$\pm$0.7}
& 78.6{\tiny$\pm$0.2}
& 80.8{\tiny$\pm$0.4}
& 81.1{\tiny$\pm$0.4} \\

MSCR
& \textbf{75.7{\tiny$\pm$1.1}}
& \textbf{86.4{\tiny$\pm$0.9}}
& \textbf{74.5{\tiny$\pm$0.2}}
& 89.3{\tiny$\pm$0.3}
& \textbf{54.4{\tiny$\pm$2.4}}
& \textbf{76.8{\tiny$\pm$0.8}}
& \textbf{80.1{\tiny$\pm$0.5}}
& \textbf{81.3{\tiny$\pm$0.5}}
& \textbf{81.8{\tiny$\pm$0.6}} \\
\bottomrule[1.5pt]
\end{tabular}
}

\vspace{1mm} 
{\footnotesize \raggedright
$\times$ method not applicable to regression tasks; \newline
$\otimes$ method not applicable to trimodal inputs\par}
\vspace{-1.5mm}
\end{table}

As shown in Table~\ref{table:main_results}, MSCR outperforms MCR in every evaluated setting. The advantage holds across task domains (audio-visual classification and multimodal sentiment analysis) and modality settings (bimodal and trimodal). The following subsections examine whether these gains come with stronger synergistic capability and how each component of MSCR contributes.

\subsection{Synergy Analysis via Conditional Recovery}
\label{sec:synergy_analysis}

\begin{figure}[!ht]
    \centering
    \begin{minipage}[t]{\textwidth}
        \centering
        \includegraphics[width=0.82\textwidth]{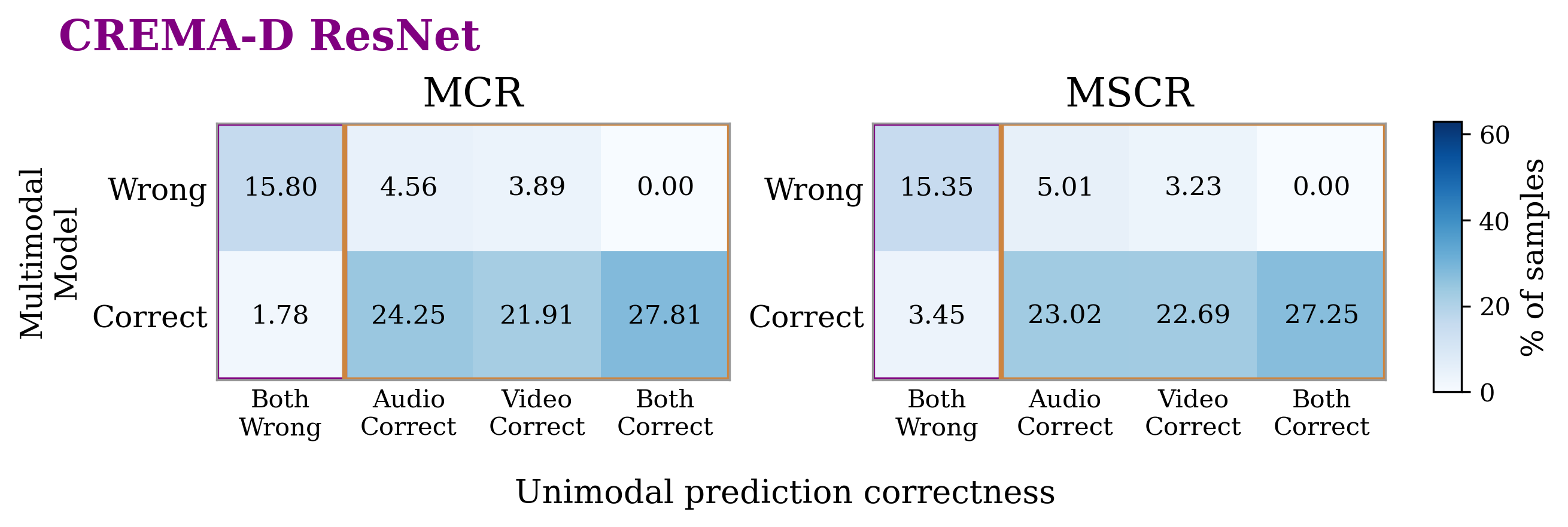}
    \end{minipage}
    \vspace{-0.15cm}
    \begin{minipage}[t]{\textwidth}
        \centering
        \includegraphics[width=0.82\textwidth]{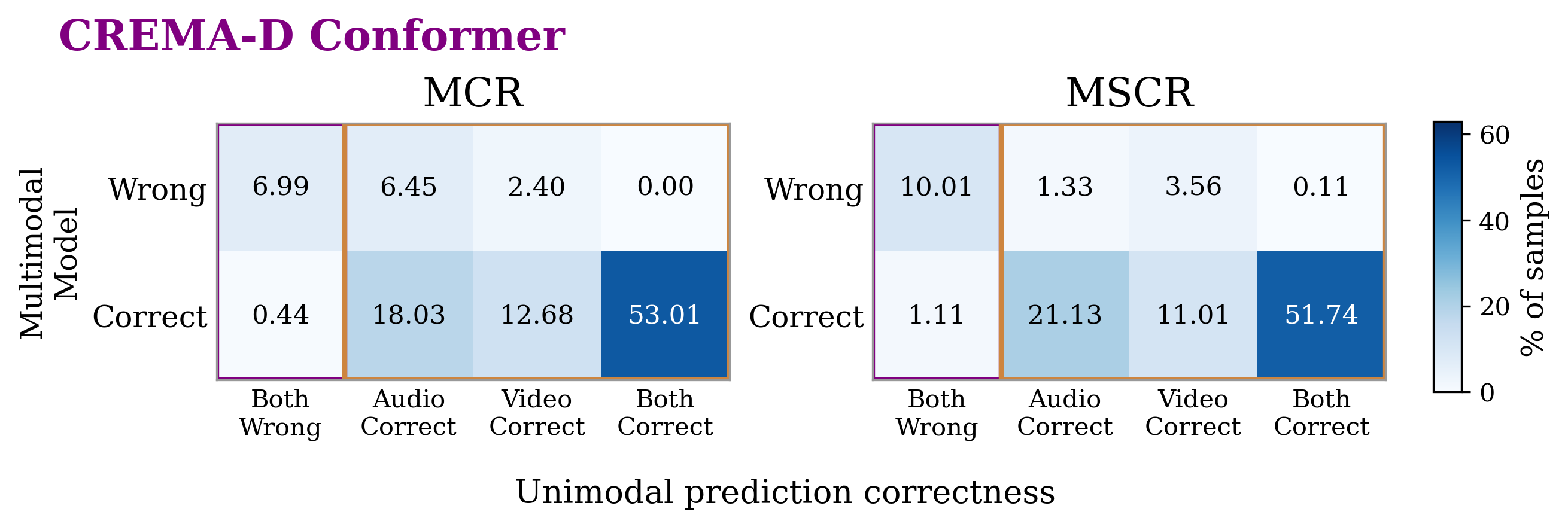}
    \end{minipage}
    \vspace{-0.2cm}
    \caption{
    Error comparison on CREMA-D with ResNet and Conformer unimodal backbones. Each matrix shows multimodal prediction performance conditioned on unimodal prediction correctness, with cells giving the percentage of samples in each category. Columns indicate whether both unimodal predictors are wrong, only audio is correct, only video is correct, or both are correct; rows indicate whether the multimodal model is correct. Results on the other datasets are given in Appendix~\ref{app:supplementary_synergy_analysis}.
    }
    \label{fig:cremad_error_analysis_main}
\vspace{-5mm}
\end{figure}

To assess synergistic capability beyond balanced modality utilization, we analyze conditional recovery rates. The idea is to examine the multimodal model on ``hard'' cases in which both unimodal predictors fail to provide sufficient evidence. Let $\hat{y}_{\mathrm{joint}}$, $\hat{y}_1$, and $\hat{y}_2$ be the predictions of the fusion network and of the two auxiliary unimodal predictors of the same trained model, and let $\mathcal{W}$ be the set of test samples on which both $\hat{y}_1$ and $\hat{y}_2$ are wrong. The conditional recovery rate is the accuracy of the joint predictor on $\mathcal{W}$:
\begin{equation}
\label{eq:recovery}
\mathrm{Recovery} = \frac{\bigl|\{a \in \mathcal{W} : \hat{y}_{\mathrm{joint}}^{(a)} = y^{(a)}\}\bigr|}{|\mathcal{W}|}.
\end{equation}
For MOSI and MOSEI, correctness follows the binary accuracy criterion. In the trimodal settings, $\mathcal{W}$ contains the samples on which all three unimodal predictions are wrong. Unlike standard task metrics, this rate counts joint predictions that succeed where neither unimodal predictor does, which gives a behavioral proxy for synergistic capability. In Figures~\ref{fig:cremad_error_analysis_main} and~\ref{fig:app_supplementary_synergy_matrices}, it is the share of the Both Wrong column that falls in the Correct row.

\begin{table}[t]
\vspace{-3mm}
\captionsetup{position=above, skip=5pt} 
\caption{Conditional recovery rates (Eq.~\eqref{eq:recovery}) of MCR and MSCR, i.e., the accuracy of the multimodal model on samples that the unimodal predictors misclassify. Higher values are better.}
\label{table:both_wrong_recovery}
\centering
\resizebox{\textwidth}{!}{
\begin{tabular}{lccccccccc}
\toprule[1.5pt]
& \multicolumn{2}{c}{CREMA-D} & \multicolumn{2}{c}{AVE} & \multicolumn{1}{c}{UCF} & \multicolumn{2}{c}{MOSI} & \multicolumn{2}{c}{MOSEI} \\
& \multicolumn{1}{c}{ResNet} & \multicolumn{1}{c}{Conformer} & \multicolumn{1}{c}{ResNet} & \multicolumn{1}{c}{Conformer} & \multicolumn{1}{c}{ResNet} & \multicolumn{4}{c}{Transformer} \\
Method & V-A & V-A & V-A & V-A & V-A & V-T & V-A-T & V-T & V-A-T \\ \midrule[1.2pt]

MCR
& 10.13\%
& 5.92\%
& 5.50\%
& 6.93\%
& 10.28\%
& 4.85\%
& 10.87\%
& 7.12\%
& 6.71\% \\

MSCR
& \textbf{18.35\%}
& \textbf{9.98\%}
& \textbf{7.07\%}
& \textbf{12.95\%}
& \textbf{15.45\%}
& \textbf{34.35\%}
& \textbf{17.44\%}
& \textbf{12.50\%}
& \textbf{13.69\%} \\
\bottomrule[1.5pt]
\end{tabular}
}
\vspace{-1mm}
\end{table}

Figure~\ref{fig:cremad_error_analysis_main} shows the error comparison on CREMA-D with ResNet~\cite{he2016deep} and Conformer~\cite{gulati2020conformer} unimodal backbones, and Table~\ref{table:both_wrong_recovery} reports the conditional recovery rates for all settings. The recovery rate of MSCR is higher than that of MCR on every dataset and backbone. The largest gain appears on MOSI in the V-T setting (from 4.85\% to 34.35\%). These results suggest that MSCR improves synergistic capability at the behavioral level: the joint predictor makes better use of latent cues that the unimodal predictors do not fully exploit. This pattern is consistent with the proficiency-aware gate, which puts more weight on samples with high unimodal prediction losses.

\subsection{Synergy Analysis via Cross-Modal Interaction Assessment}
\label{sec:intershap}

As a complement to conditional recovery, we measure model-level cross-modal interaction with Global InterSHAP~\cite{wenderoth2025measuring}. Built on the Shapley Interaction Index, it separates modality-specific contributions to the model output from interaction contributions. For sample $a$, let $\phi_{ij}^{(a)}$ ($i \neq j$) be the Shapley interaction index of modalities $i$ and $j$, and let $\phi_{ii}^{(a)}$ be the main effect of modality $i$, i.e., its Shapley value minus its interactions with the other modalities. Over $N$ evaluated samples, $\Phi_{ij}=\left|\frac{1}{N}\sum_{a=1}^{N}\phi_{ij}^{(a)}\right|$ is the absolute dataset-averaged contribution, so diagonal terms represent modality-specific effects and off-diagonal terms represent interactions. The global score is
\begin{equation}
\operatorname{Global\ InterSHAP}=\frac{\sum_{i\neq j}\Phi_{ij}}{\sum_{i,j}\Phi_{ij}}.
\label{eq:intershap_global}
\end{equation}
The score lies in $[0,1]$, is reported in percent, and measures the share of the model's output behavior that comes from cross-modal interactions. It requires neither auxiliary unimodal predictors nor ground-truth labels, so it complements the behavioral evidence from conditional recovery.

\begin{table}[t]
\centering
\caption{Global InterSHAP scores (\%, mean $\pm$ standard deviation) under identical architectures. Higher scores indicate greater relative interaction contributions.}
\label{tab:intershap}
\small
\begin{tabular}{lcc}
\toprule
Dataset / Backbone & MCR & MSCR \\
\midrule
CREMA-D / Conformer & $5.22 \pm 1.33$ & $\mathbf{10.31 \pm 1.89}$ \\
AVE / Conformer & $4.74 \pm 0.93$ & $\mathbf{9.16 \pm 1.28}$ \\
UCF / ResNet & $12.10 \pm 0.67$ & $\mathbf{18.21 \pm 1.12}$ \\
\bottomrule
\end{tabular}
\end{table}

As shown in Table~\ref{tab:intershap}, MSCR increases the relative contribution of cross-modal interactions in all three settings under identical architectures, in line with its higher conditional recovery rates.

\subsection{Synergy under Highly Interactive Fusion}
\label{sec:transformer_fusion}

The lower recovery rates of MCR might simply reflect the limited cross-modal interaction capacity of the default fusion module, which follows MCR and combines the modality representations through a single linear layer (Appendix~\ref{app:experimental_details}). To test this, we replace it with a Transformer-based fusion module implemented with two Conformer blocks, which apply self-attention to the concatenated multimodal token sequence. We then measure conditional recovery (Eq.~\eqref{eq:recovery}) for MCR and MSCR.

\begin{table}[!ht]
\vspace{-3mm}
\captionsetup{position=above, skip=5pt} 
\caption{Conditional recovery rates under Transformer-based fusion architectures. Higher values indicate a greater ability to recover correct joint predictions when both unimodal predictors fail.}
\label{tab:transformer_recovery}
\centering
\resizebox{0.50\textwidth}{!}{
\begin{tabular}{lcccc}
\toprule[1.5pt]
& \multicolumn{2}{c}{CREMA-D} & \multicolumn{1}{c}{AVE} & \multicolumn{1}{c}{UCF} \\
& \multicolumn{1}{c}{ResNet} & \multicolumn{1}{c}{Conformer} & \multicolumn{1}{c}{ResNet} & \multicolumn{1}{c}{ResNet} \\
Method & V-A & V-A & V-A & V-A \\ \midrule[1.2pt]

MCR
& 15.66\%
& 11.82\%
& 11.11\%
& 16.41\% \\

MSCR
& \textbf{21.39\%}
& \textbf{25.66\%}
& \textbf{15.13\%}
& \textbf{22.63\%} \\
\bottomrule[1.5pt]
\end{tabular}
}
\end{table}

Table~\ref{tab:transformer_recovery} shows that, even with self-attention across the concatenated multimodal tokens, MCR achieves lower conditional recovery rates than MSCR across all evaluated settings. On CREMA-D, MCR recovers 15.66\% of the both-wrong samples with ResNet encoders and 11.82\% with Conformer encoders, whereas MSCR recovers 21.39\% and 25.66\%. A more expressive fusion module therefore does not close the gap in conditional recovery between MCR and MSCR.

\subsection{Ablation Study}
\label{sec:ablation_study}

To understand the contribution of each component, we run an ablation on representative settings from the main experiments and compare four variants: the baseline without either objective (which reduces to MCR), the variant with only the anti-dominance objective $\mathcal{L}_{\mathrm{dom}}$, the variant with only the synergy objective $\mathcal{L}_{\mathrm{syn}}$, and the full MSCR with both objectives.

\begin{table}[!ht]
\vspace{-4mm}
\captionsetup{position=above, skip=5pt}
\caption{Ablation study of the anti-dominance objective $\mathcal{L}_{\mathrm{dom}}$ and the synergy objective $\mathcal{L}_{\mathrm{syn}}$ across multiple datasets, backbones, and modality settings.}
\label{table:ablation_mscr}
\centering
\small
\resizebox{\textwidth}{!}{%
\begin{tabular}{@{}ccccccccc@{}}
\toprule
$\mathcal{L}_{\mathrm{dom}}$ & $\mathcal{L}_{\mathrm{syn}}$
& CREMA-D
& AVE
& UCF
& MOSI
& MOSI
& MOSEI
& MOSEI \\
&
& Conformer
& Conformer
& ResNet
& V-T
& V-A-T
& V-T
& V-A-T \\
\midrule

&
& 85.1$\pm$0.5
& 87.4$\pm$0.2
& 52.9$\pm$2.4
& 75.3$\pm$0.7
& 78.6$\pm$0.2
& 80.8$\pm$0.4
& 81.1$\pm$0.4 \\

\checkmark
&
& 85.9$\pm$0.6
& 88.8$\pm$0.4
& 53.6$\pm$2.2
& 75.9$\pm$0.7
& 79.7$\pm$0.5
& 81.1$\pm$0.3
& 81.4$\pm$0.4 \\

& \checkmark
& 86.2$\pm$0.5
& 88.3$\pm$0.2
& 53.5$\pm$2.3
& 76.1$\pm$0.9
& 79.2$\pm$0.6
& 80.9$\pm$0.4
& 81.2$\pm$0.2 \\

\checkmark & \checkmark
& \textbf{86.4$\pm$0.9}
& \textbf{89.3$\pm$0.3}
& \textbf{54.4$\pm$2.4}
& \textbf{76.8$\pm$0.8}
& \textbf{80.1$\pm$0.5}
& \textbf{81.3$\pm$0.5}
& \textbf{81.8$\pm$0.6} \\

\bottomrule
\end{tabular}%
}
\end{table}

As shown in Table~\ref{table:ablation_mscr}, adding either $\mathcal{L}_{\mathrm{dom}}$ or $\mathcal{L}_{\mathrm{syn}}$ alone improves over the baseline in every evaluated setting, and optimizing both together gives the best results, so the two objectives are complementary.

\section{Related Work}
\label{sec:related_work}

\textbf{Imbalanced multimodal learning.} Multimodal joint training often suffers from modality laziness~\cite{wang2020makes,winterbottom2020modality,sun2021learning,du2023unimodal}, in which the model strongly prefers dominant modalities and suppresses the learning of the others. Balance-oriented strategies address this by intervening in the optimization dynamics. One category uses dynamic modulation: OGM~\cite{peng2022balanced}, AGM~\cite{li2023boosting}, and MMPareto~\cite{wei2024mmpareto} adjust gradients or loss weights to penalize the dominant modality~\cite{vielzeuf2018centralnet,xu2023mmcosine,kontras2024improving}, and MSLR~\cite{yao2022modality} adaptively scales learning rates according to performance discrepancies. Another category decouples training: D\&R~\cite{wei2024diagnosing}, MLA~\cite{zhang2024multimodal}, and ReconBoost~\cite{hua2024reconboost} isolate unimodal updates to avoid interference between modalities. MCR~\cite{kontras2025balancing} recently introduced an information-decomposition framework to mitigate dominance, and IBML~\cite{qin2026deep} addresses modality imbalance by adaptively balancing the retention of complementary information across modalities. These approaches mainly aim to equalize the contributions of individual modalities and do not require the fusion network to produce collaborative gains, so a balanced model may still fail to exploit synergistic information.

\textbf{Information-theoretic multimodal learning.} Information bottleneck methods study the trade-off between compressing and preserving information~\cite{hu2024survey}. MIGE~\cite{wen2020mutual} directly estimates mutual information gradients for representation learning, and MVEB~\cite{wen2024mveb} learns minimal sufficient representations by jointly maximizing the agreement between augmented views and the differential entropy of the embedding distribution. For multiview clustering, S2WIB~\cite{lou2026structure} combines self-supervised view weighting with complementary information and cluster-structure consistency, and PIB~\cite{hu2024propagating} couples feature subspace learning and clustering through iterative information propagation. Beyond preserving information, one can ask whether predictive information is available from individual modalities or requires joint observation. PID~\cite{williams2010nonnegative,bertschinger2014quantifying} formalizes this distinction through redundant, unique, and synergistic components. Recent studies~\cite{xin2025i2moe,xiu2026comprehensive} argue that the main benefit of multimodal learning comes from synergistic information, that is, cues that emerge only from joint interaction and cannot be recovered from unimodal evidence alone~\cite{liang2023factorized}. To capture such interactions, FactorCL~\cite{liang2023factorized} and CoMM~\cite{dufumier2025align} model interaction patterns with multimodal augmentations and contrastive objectives, and masking strategies have been used to uncover hidden synergistic semantics~\cite{wen2025infmasking}. Our work targets supervised multimodal fusion: we introduce a fusion-level regularizer that encourages joint predictive gain beyond unimodal aggregation and can complement existing modality-balancing methods.

\section{Conclusion}
\label{sec:conclusion}

We proposed MSCR, a fusion-level regularizer that addresses modality imbalance and explicitly encourages the discovery of synergistic information. MSCR acts only on the fusion bottleneck and does not interfere with the unimodal encoders. {It penalizes large perturbation sensitivities to regularize modality reliance and encourages the joint prediction to outperform an additive unimodal reference. A proficiency-aware gate adjusts the regularization strength for each sample.} Experiments on five benchmarks show that MSCR consistently improves downstream performance over MCR and increases synergistic capability, as measured by conditional recovery and InterSHAP. A limitation is that the synergy objective uses an additive unimodal reference as a tractable approximation, which may not capture every form of multimodal interaction. A natural next step is a more exact synergy measure with theoretical bounds.

\section*{Acknowledgments}

This work was supported by the Sichuan Provincial Innovation Group Project (Grant No. 2024NSFTD0054), the Major Science and Technology Special Project of the Sichuan Provincial Department of Science and Technology (Grant No. 2024ZDZX0002), Fundamental Research Funds for the Central Universities (JBK202511081), the National Natural Science Foundation of China (No. U24A20323, No. 62502397).
We thank anonymous reviewers for their insightful feedback and Konstantinos Kontras, the first author of MCR~\cite{kontras2025balancing}, for his contributions to open sourcing the code.

\newpage
\bibliography{reference.bib}
\bibliographystyle{plain}

\clearpage
\appendix
\newcommand{\reddashedline}{\textcolor{red}{\rule[0.5ex]{0.3em}{1pt}\hspace{0.2em}\rule[0.5ex]{0.3em}{1pt}\hspace{0.2em}\rule[0.5ex]{0.3em}{1pt}\hspace{0.2em}\rule[0.5ex]{0.3em}{1pt}}}

\begin{center}
    \LARGE \textbf{Appendix}
\end{center}

\section{Experimental details}
\label{app:experimental_details}

This section gives the details needed to reproduce our experiments. For all datasets we follow the evaluation protocol of MCR~\cite{kontras2025balancing} and use the same data splits, modality-specific encoders, fusion backbones, optimizer settings, learning-rate schedules, batch sizes, and training epochs. The checkpoint with the best validation performance is evaluated on the test set. For CREMA-D~\cite{cao2014cremad} and UCF~\cite{soomro2012ucf101} we use a three-fold evaluation protocol and report the mean and standard deviation across folds. For AVE~\cite{tian2018audio}, MOSI~\cite{zadeh2016mosi}, and MOSEI~\cite{zadeh2018multimodal} we report the mean and standard deviation over three independent runs with different random seeds.

For the audio-visual classification datasets (CREMA-D, AVE, and UCF) we report classification accuracy. For the multimodal sentiment analysis datasets (MOSI and MOSEI) we follow the regression protocol and metrics of MCR. Experiments with pre-trained unimodal encoders use the same initialization and fine-tuning strategy as MCR. In the regression tasks on CMU-MOSI and CMU-MOSEI, the unimodal and joint predictors produce scalar outputs, so we compute modality sensitivities with squared error instead of JSD and use MSE for the joint, reference, and unimodal losses in the synergy objective and the adaptive gate. The reference is the mean of the unimodal outputs.

Following MCR, our default fusion module sums modality-specific linear projections with a shared bias, which is equivalent to applying a single linear prediction layer to the concatenated modality representations. For the fusion replacement experiment in Section~\ref{sec:transformer_fusion}, the Transformer-based fusion module consists of two Conformer blocks with input and hidden dimensions of 512 and eight attention heads, and applies self-attention to the concatenated multimodal token sequence.

\begin{table}[ht]
    \centering
    \vspace{-2.5mm}
    \caption{Hyperparameters for MSCR. $\lambda_{\mathrm{dom}}$ and $\lambda_{\mathrm{syn}}$ denote the weights of the anti-dominance objective and the synergy objective. $m$ is the margin in the synergy objective. $\tau_{\mathrm{dom}}$ is the smoothmax temperature. $\kappa$ and $\tau$ are the sharpness and threshold of the adaptive modulation gate.}
    \label{table:smcr_hyperparameters}
    \begin{tabular}{lcccccc}
        \toprule
        \textit{dataset} & $\lambda_{\mathrm{dom}}$ & $\lambda_{\mathrm{syn}}$ & $m$ & $\tau_{\mathrm{dom}}$ & $\kappa$ & $\tau$ \\
        \midrule
        \textit{CREMA-D ResNet V-A} & 0.02 & 0.03 & 0.05 & 10.0 & 5.0 & 1.5 \\
        \textit{CREMA-D Conformer V-A} & 0.05 & 0.03 & 0.05 & 10.0 & 5.0 & 1.2 \\
        \textit{AVE ResNet V-A} & 0.02 & 0.03 & 0.05 & 10.0 & 5.0 & 1.5 \\
        \textit{AVE Conformer V-A} & 0.05 & 0.03 & 0.05 & 10.0 & 5.0 & 1.2 \\
        \textit{UCF ResNet V-A} & 0.02 & 0.03 & 0.05 & 10.0 & 5.0 & 1.5 \\
        \textit{MOSI Transformer V-T} & 0.05 & 0.03 & 0.04 & 10.0 & 5.0 & 1.2 \\
        \textit{MOSI Transformer V-A-T} & 0.05 & 0.03 & 0.04 & 10.0 & 5.0 & 1.2 \\
        \textit{MOSEI Transformer V-T} & 0.05 & 0.03 & 0.04 & 10.0 & 5.0 & 1.2 \\
        \textit{MOSEI Transformer V-A-T} & 0.05 & 0.03 & 0.04 & 10.0 & 5.0 & 1.2 \\
        \bottomrule
    \end{tabular}
\end{table}

Table~\ref{table:smcr_hyperparameters} lists the MSCR-specific hyperparameters (anti-dominance weight, synergy weight, margin, smoothmax temperature, gate sharpness, and gate threshold) for each setting. $\lambda_{\mathrm{syn}}$, $\tau_{\mathrm{dom}}$, and $\kappa$ are shared across all settings; $\lambda_{\mathrm{dom}}$, $m$, and $\tau$ differ only between backbone families. MSCR is activated during epochs $[T_s, T_e]=[10,30]$ for models with ResNet or Transformer encoders and $[5,15]$ for models with Conformer encoders (Algorithm~\ref{alg:mscr}), with both endpoints included. All experiments are conducted on a single NVIDIA RTX 4090 GPU with 24 GB of memory. A single training run takes approximately 0.5--1.5 hours on CMU-MOSI and CMU-MOSEI, and around 5 hours on CREMA-D, AVE, and UCF, depending on the experimental setting. The code, configuration files, and running scripts are publicly available at \url{https://github.com/chenxinyu991-collab/MSCR}.

\section{Dataset and backbone details}
\label{app:dataset_backbone_details}

\subsection{Datasets}
\label{app:datasets}

Following the experimental protocols of MCR~\cite{kontras2025balancing}, we describe each dataset and its preprocessing below.

\textbf{CREMA-D}~\cite{cao2014cremad} is an audio-visual emotion recognition dataset of video clips from 91 actors, each expressing six emotions: Happy, Sad, Anger, Fear, Disgust, and Neutral. Following the preprocessing in~\cite{wang2020makes, peng2022balanced, li2023boosting}, video frames are sampled at 1 fps (3 consecutive frames) and audio at 22 kHz (3 seconds). Log-Mel spectrograms are computed via STFT with a window size of 512 and a step size of 353. For the stronger models we follow~\cite{goncalves2023versatile} and use 16 kHz audio and pre-computed facial features. Our data split also follows~\cite{goncalves2023versatile}, so no actor appears in more than one of the training, validation, and test sets.

\textbf{AVE}~\cite{tian2018audio} contains 4,143 videos in 28 event categories, each with temporally labeled audio-visual events of at least 2 seconds. Following~\cite{wang2020makes}, we sample 4 video frames at 1 fps and resample audio at 16 kHz with the same STFT settings as CREMA-D. We use the official predefined splits and report the standard deviation across three random seeds on the test set.

\textbf{UCF101}~\cite{soomro2012ucf101} is a large-scale action recognition dataset. Following the multimodal protocol in~\cite{wang2020makes}, we use the subset of 51 action categories that contain both audio and video streams. Preprocessing is the same as for AVE. We use the standard 3-fold split of the original dataset and report the standard deviation across these folds.

\textbf{CMU-MOSI}~\cite{zadeh2016mosi} and \textbf{CMU-MOSEI}~\cite{zadeh2018multimodal} are multimodal sentiment analysis benchmarks in which each segment is annotated with a sentiment score from $-3$ to $3$. Standard metrics for these datasets~\cite{tsai2019multimodal} include 7-class accuracy, binary accuracy, F1 score, mean absolute error, and correlation; following MCR, we report binary accuracy. We use the aligned versions of both datasets provided by MultiBench~\cite{liang2021multibench}. Results are the mean and standard deviation across three random seeds on the test set.

\subsection{Models: Backbone Unimodal Encoders}
\label{app:backbone_unimodal_encoders}

Following previous work~\cite{li2023boosting, peng2022balanced, xu2023mmcosine, wang2020makes}, our standard experiments use ResNet-18~\cite{he2016deep} as the unimodal encoder for both video and audio in CREMA-D, AVE, and UCF101. These encoders are randomly initialized and use adaptive pooling to handle different input dimensions. For CMU-MOSI and CMU-MOSEI we use a 5-layer Transformer~\cite{vaswani2017attention} backbone, following~\cite{liang2024foundations, liang2023factorized}.

To test whether MSCR remains effective with encoders of different capacity, we also evaluate stronger, pre-trained unimodal encoders.

\textbf{CREMA-D Conformer setting.} Following~\cite{goncalves2023versatile}, we use the first 12 layers of Wav2Vec2~\cite{baevski2020wav2vec, wolf2020transformers}, initialized with self-supervised pre-trained weights for speech recognition and fine-tuned during training. For video, MTCNN~\cite{zhang2016joint} extracts face bounding boxes and a frozen EfficientNet-B2~\cite{tan2019efficientnet} serves as the feature extractor. A 5-layer Conformer~\cite{gulati2020conformer} trained from scratch further processes the audio and facial features. The full model has about 183M parameters. We refer to this configuration as the ``Conformer'' setting.

\textbf{AVE Conformer setting.} We use a similar architecture for AVE. The visual branch is a ViViT model~\cite{arnab2021vivit} pre-trained on Kinetics~\cite{kay2017kinetics}, and the audio branch is HuBERT~\cite{hsu2021hubert} pre-trained on AudioSet~\cite{gemmeke2017audio}. A 5-layer Conformer~\cite{gulati2020conformer} processes both representations, for about 232M parameters in total. With these encoders, the pre-trained knowledge covers non-speech audio events and complex temporal patterns in video.

\section{Supplementary synergy analysis}
\label{app:supplementary_synergy_analysis}

This section gives the remaining error comparison matrices for the conditional recovery analysis in Section~\ref{sec:synergy_analysis}, covering AVE~\cite{tian2018audio} with ResNet and Conformer~\cite{gulati2020conformer} backbones, MOSI~\cite{zadeh2016mosi} and MOSEI~\cite{zadeh2018multimodal} in the V-T setting, and UCF~\cite{soomro2012ucf101} with the ResNet backbone. The error comparison matrices are computed on the test split from a single run for each setting, using seed 42 for datasets evaluated across random seeds and fold 0 for datasets evaluated across cross-validation folds. In contrast, Global InterSHAP scores are reported as mean $\pm$ standard deviation across random seeds or cross-validation folds, following the evaluation protocol of each dataset. Each matrix breaks down the joint predictions by the correctness of the individual unimodal predictions. The ``Both Wrong'' cases, in which both unimodal predictors fail, provide behavioral evidence of the joint predictor's synergistic capability.

\begin{figure}[!ht]
    \centering
    \includegraphics[width=0.78\textwidth]{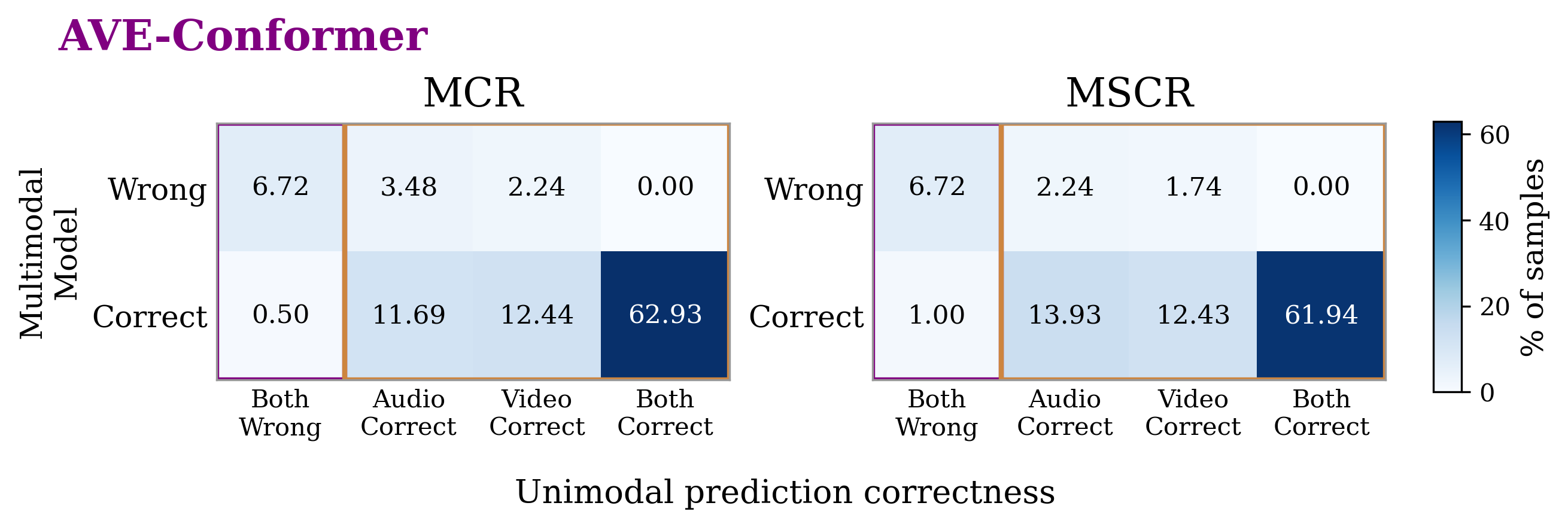} \\[-0.1cm]
    
    \includegraphics[width=0.78\textwidth]{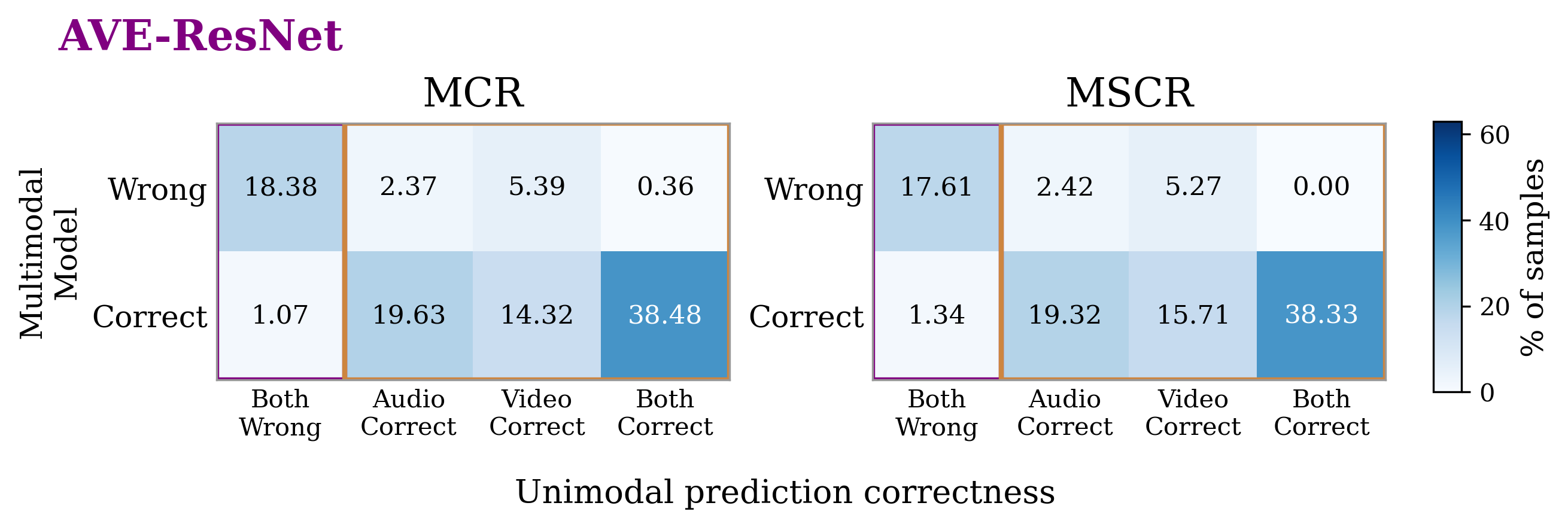} \\[-0.1cm]
    
    \includegraphics[width=0.78\textwidth]{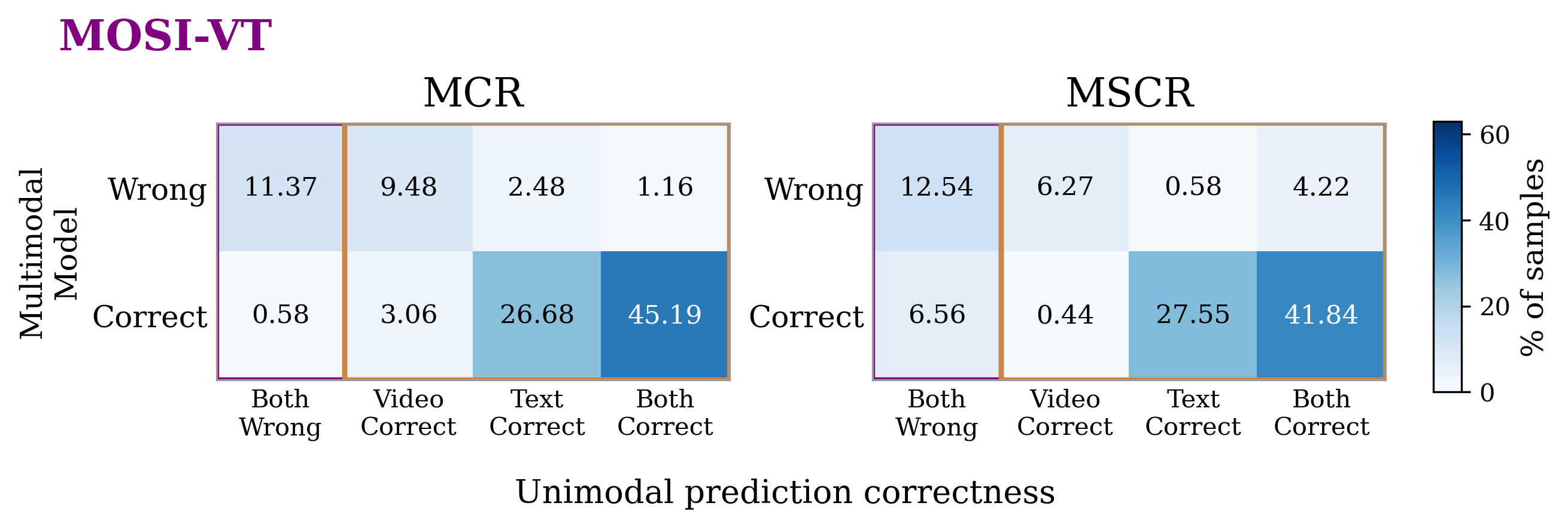} \\[-0.1cm]
    
    \includegraphics[width=0.78\textwidth]{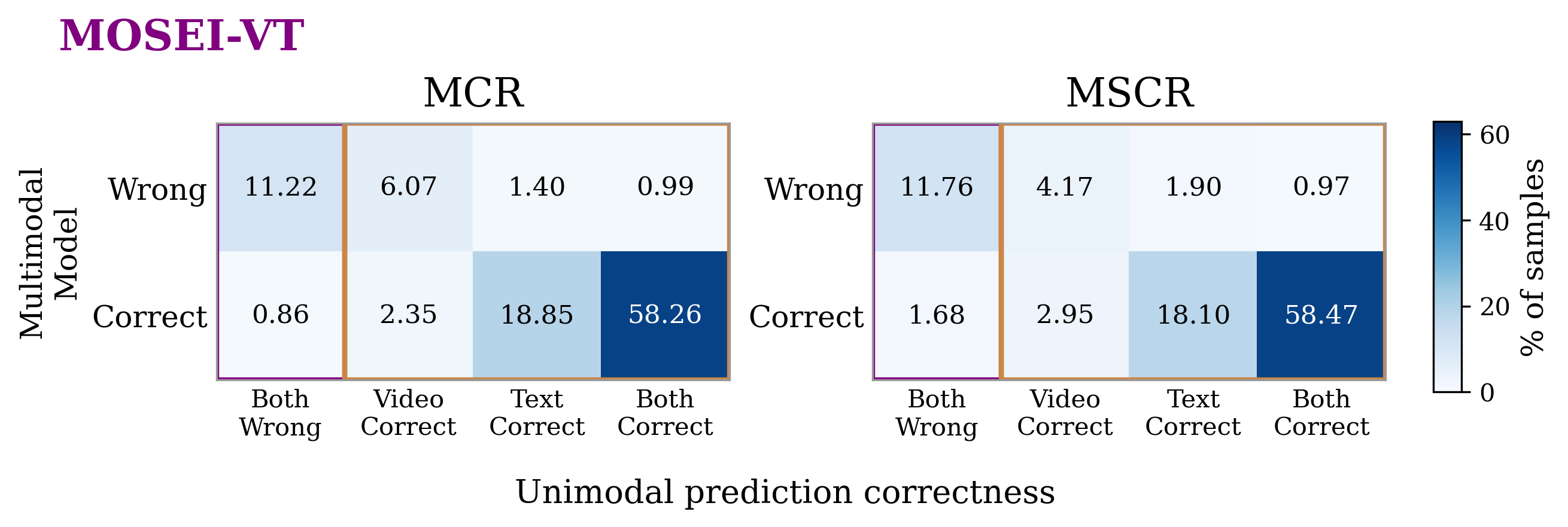} \\[-0.1cm]
    
    \includegraphics[width=0.78\textwidth]{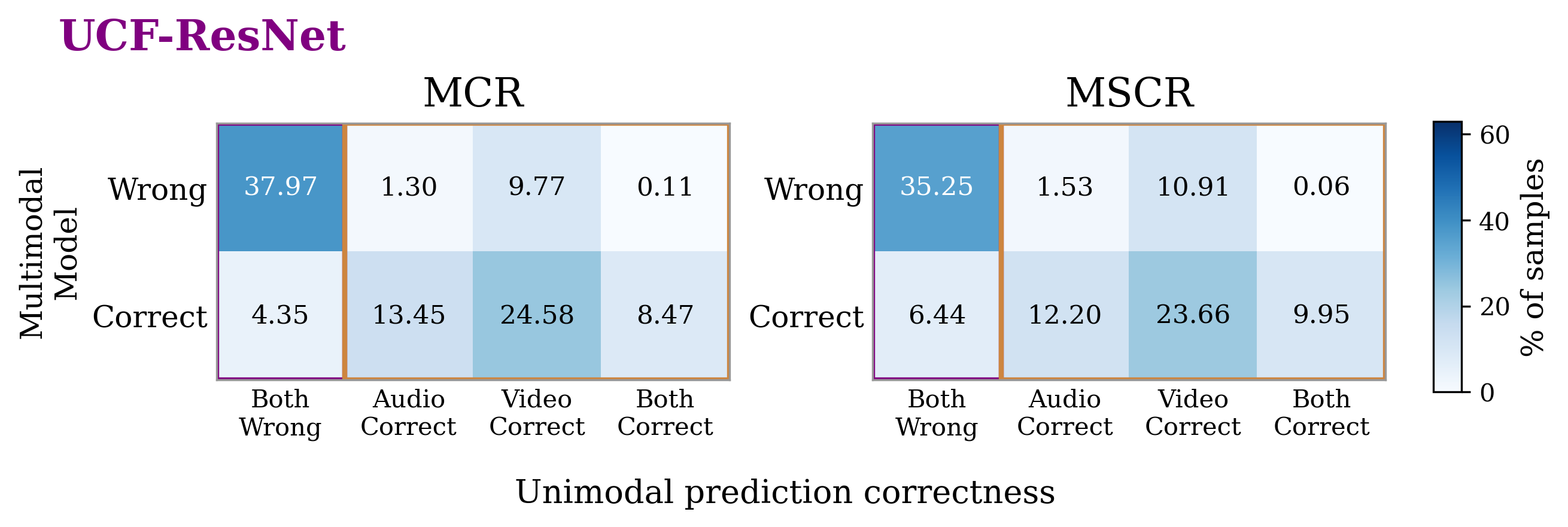} \\[-0.1cm]
    
    \caption{
    Error comparison matrices for additional datasets and backbones. Each matrix groups the multimodal predictions by the correctness of the corresponding unimodal predictions. Compared with MCR, MSCR achieves higher conditional recovery in the ``Both Wrong'' cases (unimodal predictors fail but the multimodal model is correct), which indicates a stronger ability to extract synergistic information when unimodal evidence is insufficient.
    }
    \label{fig:app_supplementary_synergy_matrices}
\end{figure}

Consistent with the main text, MSCR recovers a larger share of the Both Wrong samples than MCR~\cite{kontras2025balancing} in all five settings of Figure~\ref{fig:app_supplementary_synergy_matrices}.

\section{Independence and Compatibility of MSCR}
\label{app:mscr_compatibility}

Our main experiments combine MSCR with MCR~\cite{kontras2025balancing}, but the regularizer does not depend on MCR-specific objectives. To test this, we remove all MCR regularization terms and keep only the supervised task loss and MSCR (\textit{MSCR-only}). We also add MSCR to another modality-balancing method, OGM~\cite{peng2022balanced}. Within each setting, all variants share the same modality encoders and fusion architecture. Table~\ref{tab:mscr_compatibility} shows that MSCR-only performs comparably to MCR and that combining MSCR with MCR further improves accuracy. Adding MSCR to OGM improves performance in all three settings.

\begin{table}[ht]
\centering
\caption{Independence and compatibility of MSCR. MSCR-only removes all MCR-specific regularization terms. Results are accuracy (\%), reported as mean $\pm$ standard deviation. OGM results are taken from~\cite{kontras2025balancing}; all other results are obtained from our own experiments.}
\label{tab:mscr_compatibility}
\small
\setlength{\tabcolsep}{5pt}
\begin{tabular}{lccc}
\toprule
Method & \shortstack{CREMA-D\\Conformer} & \shortstack{AVE\\Conformer} & \shortstack{UCF\\ResNet} \\
\midrule
OGM & $82.4 \pm 1.0$ & $79.7 \pm 1.4$ & $51.8 \pm 1.9$ \\
OGM + MSCR & $83.5 \pm 1.2$ & $81.1 \pm 1.0$ & $52.9 \pm 1.7$ \\
\midrule
MCR & $85.1 \pm 0.5$ & $87.4 \pm 0.2$ & $52.9 \pm 2.4$ \\
MSCR-only (w/o MCR) & $85.3 \pm 0.7$ & $86.8 \pm 0.4$ & $52.7 \pm 2.1$ \\
MCR + MSCR (full) & $\mathbf{86.4 \pm 0.9}$ & $\mathbf{89.3 \pm 0.3}$ & $\mathbf{54.4 \pm 2.4}$ \\
\bottomrule
\end{tabular}
\end{table}

\section{Third-Party Assets and Licenses}
\label{app:asset_licenses}

We summarize the licensing information for third-party assets below. Resource links and available license notices will be documented in our code repository.

\paragraph{Datasets.}
CREMA-D is distributed under the Open Database License (ODbL 1.0), with individual database contents licensed under the Database Contents License (DbCL 1.0). For CMU-MOSI and CMU-MOSEI, we use the preprocessed features provided through MultiBench. MultiBench and the CMU Multimodal SDK provide MIT-licensed code; this software license should be distinguished from rights in the underlying source videos. The official AVE and UCF101 pages do not state a dataset license.

\paragraph{Pretrained models and code.}
The model cards for \texttt{facebook/wav2vec2-large-robust}, \texttt{ALM/hubert-base-audioset}, and \texttt{google/vivit-b-16x2-kinetics400} specify Apache 2.0, CC BY-NC-SA 4.0, and MIT licenses, respectively. The Conformer implementation in the MCR codebase includes an Apache 2.0 license notice. We acknowledge the MCR and InterSHAP implementations. The README of the MCR repository indicates the MIT license, and the InterSHAP repository does not state a license.

\section{Pseudo-code}
\label{app:pseudo_code}

Algorithm~\ref{alg:mscr} gives the training procedure of MSCR for bimodal classification. For each mini-batch, the model extracts modality-specific representations and computes unimodal and fused predictions. Within the MSCR activation interval, it builds modality-perturbed fused predictions to estimate modality dominance, forms an additive unimodal reference for synergy enhancement, and uses an adaptive gate to scale the regularization. Outside the activation interval, training uses the supervised task loss and MCR regularization.
\clearpage
\begin{algorithm}[H]
\caption{Multimodal Training with MSCR}
\label{alg:mscr}
\begin{algorithmic}[1]
\STATE \textbf{Input:} Training dataset $\mathcal{D}$; modality encoders
$f_1,f_2$; unimodal predictors $q_1,q_2$; fusion network $f_c$;
MSCR weights $\lambda_{\mathrm{dom}},\lambda_{\mathrm{syn}}$;
margin $m$; smoothmax temperature $\tau_{\mathrm{dom}}$;
gate parameters $\kappa,\tau$; activation interval $[T_s,T_e]$.
\STATE \textbf{Notation:} $\operatorname{sg}(\cdot)$ stops gradients;
$f_c^{\mathrm{fix}}$ uses fixed fusion parameters while retaining
gradients with respect to its inputs.
\FOR{each training epoch $t$}
    \FOR{each mini-batch $\mathcal{B}=\{(\mathbf{x}_1,\mathbf{x}_2,y)\}$}
        \STATE Extract representations
        $\mathbf{z}_1=f_1(\mathbf{x}_1)$,
        $\mathbf{z}_2=f_2(\mathbf{x}_2)$.
        \STATE Compute unimodal predictions
        $p_1=q_1(\mathbf{z}_1)$,
        $p_2=q_2(\mathbf{z}_2)$,
        and joint prediction
        $p_{\mathrm{joint}}=f_c(\mathbf{z}_1,\mathbf{z}_2)$.
        \STATE Compute the supervised task loss
        $\mathcal{L}_{\mathrm{task}}$ using
        $p_{\mathrm{joint}},p_1,p_2$.

        \STATE \textit{MCR branch: regularize modality encoders.}
        \STATE Compute $\mathcal{L}_{\mathrm{MCR}}$ according to
        Eq.~\eqref{eq:mcr_loss}, using $f_c^{\mathrm{fix}}$ for MIPD
        and retaining gradients through $\mathbf{z}_1,\mathbf{z}_2$.
        \STATE MCR regularizers do not update $f_c$;
        the CEB term also updates its reconstruction head.

        \IF{$T_s \leq t \leq T_e$}
            \STATE \textit{MSCR branch: regularize the fusion network.}
            \STATE Detach representations:
            $\bar{\mathbf{z}}_1=\operatorname{sg}(\mathbf{z}_1)$,
            $\bar{\mathbf{z}}_2=\operatorname{sg}(\mathbf{z}_2)$.
            \STATE Compute
            $\bar{p}_{\mathrm{joint}}
            =f_c(\bar{\mathbf{z}}_1,\bar{\mathbf{z}}_2)$.
            \STATE Independently permute
            $\bar{\mathbf{z}}_1,\bar{\mathbf{z}}_2$
            along the batch dimension to obtain
            $\tilde{\bar{\mathbf{z}}}_1,
             \tilde{\bar{\mathbf{z}}}_2$.
            \STATE Compute
            $p_{\mathrm{shuf}1}
            =f_c(\tilde{\bar{\mathbf{z}}}_1,\bar{\mathbf{z}}_2)$,
            $p_{\mathrm{shuf}2}
            =f_c(\bar{\mathbf{z}}_1,\tilde{\bar{\mathbf{z}}}_2)$.
            \STATE Compute per-sample sensitivities
            $s_i=\operatorname{JSD}
            (\bar{p}_{\mathrm{joint}}\|p_{\mathrm{shuf}i})$,
            $i\in\{1,2\}$.
            \STATE Compute
            $\ell_{\mathrm{dom}}
            =\frac{1}{\tau_{\mathrm{dom}}}
            \log\!\left(
            e^{\tau_{\mathrm{dom}}s_1}
            +e^{\tau_{\mathrm{dom}}s_2}\right)$.

            \STATE Construct the detached reference
            $p_{\mathrm{ref}}
            =\operatorname{sg}\!\left(\frac{p_1+p_2}{2}\right)$.
            \STATE Compute
            $\ell_{\mathrm{joint}}
            =\mathcal{L}_{\mathrm{CE}}(\bar{p}_{\mathrm{joint}},y)$,
            $\ell_{\mathrm{ref}}
            =\mathcal{L}_{\mathrm{CE}}(p_{\mathrm{ref}},y)$.
            \STATE Compute the per-sample synergy loss
            $\ell_{\mathrm{syn}}
            =\max(0,m+\ell_{\mathrm{joint}}-\ell_{\mathrm{ref}})$.

            \STATE Compute detached unimodal losses
            $\ell_i=\operatorname{sg}
            (\mathcal{L}_{\mathrm{CE}}(p_i,y))$,
            $i\in\{1,2\}$.
            \STATE Compute the per-sample gate
            $g=\sigma\!\left(
            \kappa[\min(\ell_1,\ell_2)-\tau]\right)$.
            \STATE Aggregate over the mini-batch:
            $\mathcal{L}_{\mathrm{MSCR}}
            =\frac{1}{|\mathcal{B}|}
            \sum_{b\in\mathcal{B}}
            g^{(b)}
            \left(
            \lambda_{\mathrm{dom}}\ell_{\mathrm{dom}}^{(b)}
            +\lambda_{\mathrm{syn}}\ell_{\mathrm{syn}}^{(b)}
            \right)$.
        \ELSE
            \STATE Set $\mathcal{L}_{\mathrm{MSCR}}=0$.
        \ENDIF

        \STATE Set
        $\mathcal{L}_{\mathrm{total}}
        =\mathcal{L}_{\mathrm{task}}
        +\mathcal{L}_{\mathrm{MCR}}
        +\mathcal{L}_{\mathrm{MSCR}}$.
        \STATE Backpropagate $\mathcal{L}_{\mathrm{total}}$
        through the designated gradient paths and update parameters.
    \ENDFOR
\ENDFOR
\end{algorithmic}
\end{algorithm}

\end{document}